\pdfoutput=1
\documentclass[11pt,letterpaper]{article}
\usepackage{acl2015}

\usepackage{times}

\usepackage{url}
\usepackage{latexsym}
\usepackage{amsfonts, amsmath, amsthm}
\usepackage{multirow}
\usepackage{graphicx}
\usepackage{color}
\usepackage{enumerate}
\usepackage{enumitem}
\usepackage{comment}
\usepackage{bbm}
\usepackage{underscore}
\usepackage{stmaryrd}
\usepackage{bbold}
\usepackage{fixltx2e}
\usepackage{booktabs}

\newcommand{\ourtitle}{Text to 3D Scene Generation with Rich Lexical Grounding}

\newcommand{\lrtest}[4]{$\chi^2(#1,N=#2)=#3, p<#4$}

\makeatletter 
\newcommand{\@BIBLABEL}{\@emptybiblabel} 
\newcommand{\@emptybiblabel}[1]{} 
\makeatother
\usepackage[draft,pdfencoding=auto]{hyperref}
\hypersetup{
pdftitle={\ourtitle},
pdfauthor={Angel Chang, Will Monroe, Manolis Savva, Christopher Potts and Christopher D. Manning},
pdfkeywords={grounding, 3D scenes, text to 3D scene generation},
bookmarksnumbered,
pdfstartview={FitH},
colorlinks,
citecolor=black,
filecolor=black,
linkcolor=black,
urlcolor=black,
breaklinks=true,
}

\newcommand{\cat}[1]{{\footnotesize\texttt{#1}}}

\newcommand{\term}{\textit}
\newcommand{\lang}{\textit}

\DeclareMathOperator*{\argmax}{arg\,max}

\title{\ourtitle}

\author{
  Angel Chang\thanks{\;\;The first two authors are listed in alphabetical order.},\quad Will Monroe\footnotemark[1],\quad Manolis Savva,\\
  {\bf Christopher Potts} \and {\bf Christopher D. Manning} \\
  Stanford University, Stanford, CA 94305 \\
  {\tt \{angelx,wmonroe4,msavva\}@cs.stanford.edu}, \\
  {\tt \{cgpotts,manning\}@stanford.edu} \\
}

\date{}

\begin{document}
\maketitle
\begin{abstract}
The ability to map descriptions of scenes to 3D geometric representations has many applications in areas such as art, education, and robotics. However, prior work on the \emph{text to 3D scene generation} task has used manually specified object categories and language that identifies them.  We introduce a dataset of 3D scenes annotated with natural language descriptions and learn from this data how to ground textual descriptions to physical objects.  Our method successfully grounds a variety of lexical terms to concrete referents, and we show quantitatively that our method improves 3D scene generation over previous work using purely rule-based methods.  We evaluate the fidelity and plausibility of 3D scenes generated with our grounding approach through human judgments.  To ease evaluation on this task, we also introduce an automated metric that strongly correlates with human judgments.
\end{abstract}

\section{Introduction}

\begin{figure}
  \includegraphics[width=\columnwidth]{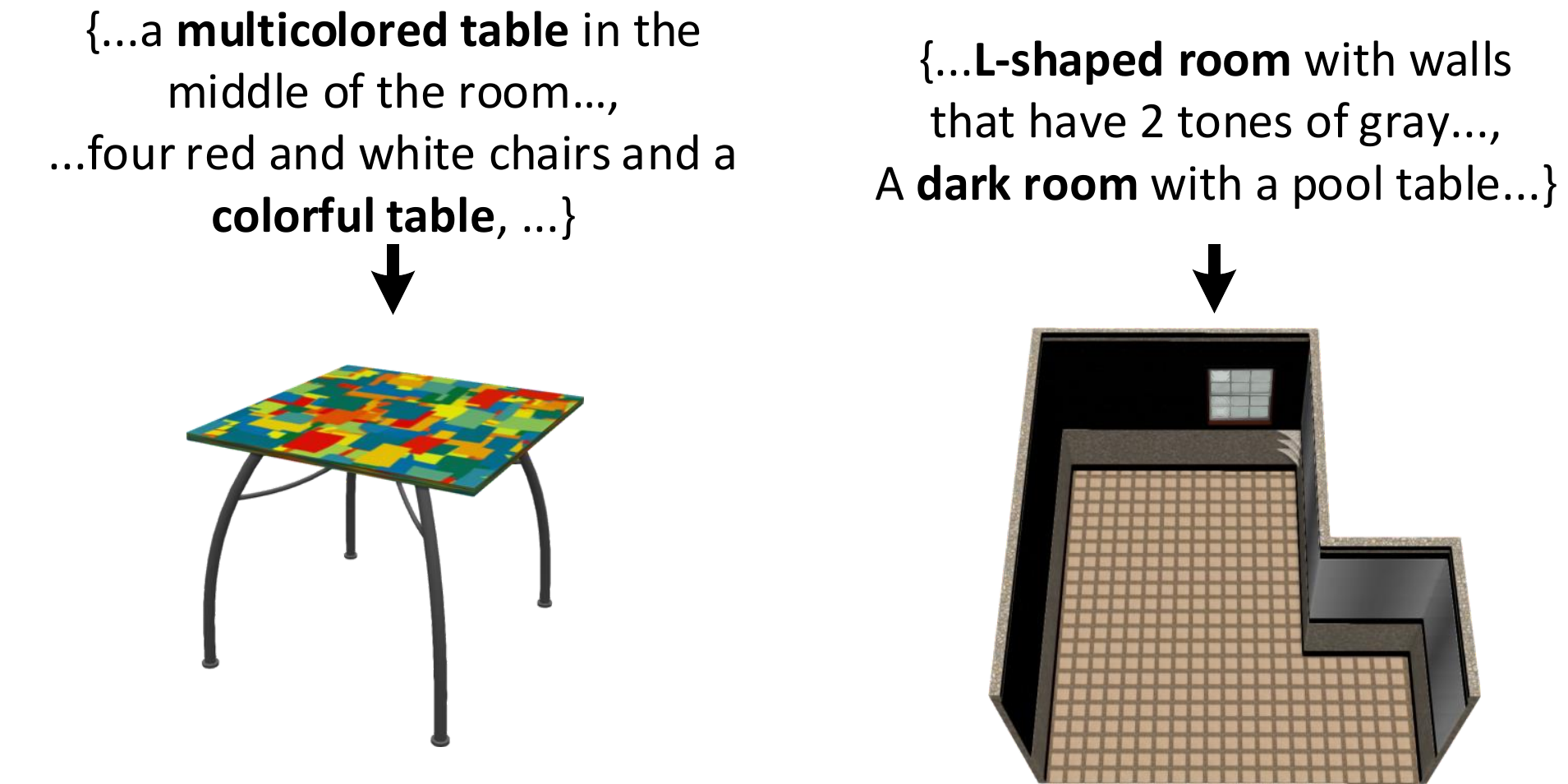}
  \caption{We learn how to ground references such as ``L-shaped room'' to 3D models in a paired corpus of 3D scenes and natural language descriptions. Sentence fragments in bold were identified as high-weighted references to the shown objects.}
  \label{fig:grounding}
\end{figure}

We examine the task of \emph{text to 3D scene generation}.  The ability to map descriptions of scenes to 3D geometric representations has a wide variety of applications; many creative industries use 3D scenes. Robotics applications need to interpret commands referring to real-world environments, and the ability to visualize scenarios given high-level descriptions is of great practical use in educational tools.  Unfortunately, 3D scene design user interfaces are prohibitively complex for novice users.  Prior work has shown the task remains challenging and time intensive for non-experts, even with simplified interfaces~\cite{savva2014sizes}.
\newpage
Language offers a convenient way for designers to express their creative goals. Systems that can interpret natural descriptions to build a visual representation allow non-experts to visually express their thoughts with language, as was demonstrated by WordsEye, a pioneering work in text to 3D scene generation~\cite{coyne2001wordseye}.

\begin{figure*}
  \includegraphics[width=\textwidth]{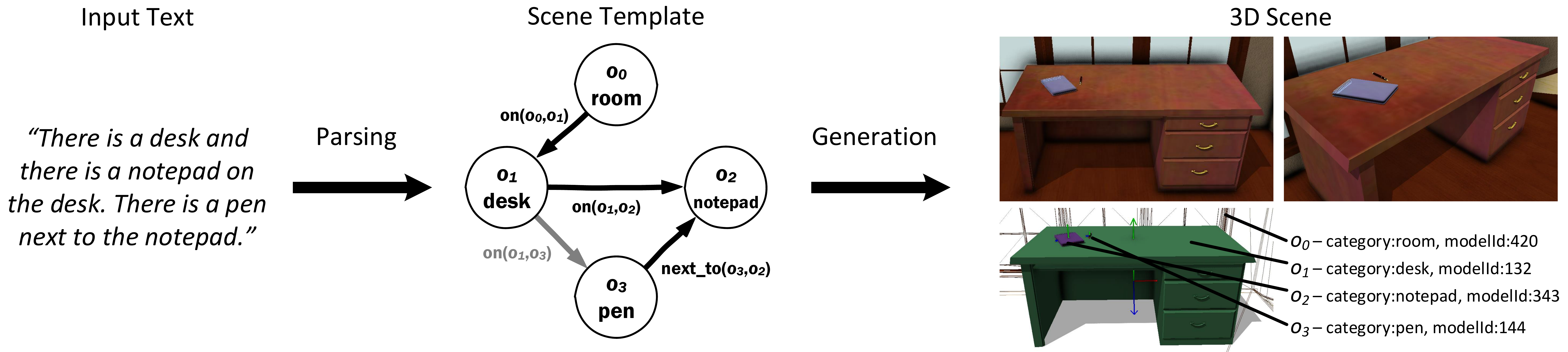} %
  \caption{Illustration of the text to 3D scene generation pipeline. The input is text describing a scene (left), which we parse into an abstract scene template representation capturing objects and relations (middle). The scene template is then used to generate a concrete 3D scene visualizing the input description (right). The 3D scene is constructed by retrieving and arranging appropriate 3D models.}
  \label{fig:text2scene-pipeline}
\end{figure*}

WordsEye and other prior work in this area~\cite{seversky2006real,chang2014spatial} used manually chosen mappings between language and objects in scenes.  To our knowledge, we present the first 3D scene generation approach that learns from data how to map textual terms to objects.  First, we collect a dataset of 3D scenes along with textual descriptions by people, which we contribute to the community.  We then train a classifier on a scene discrimination task and extract high-weight features that ground lexical terms to 3D models.  We integrate our learned lexical groundings with a rule-based scene generation approach, and we show through a human-judgment evaluation that the combination outperforms both approaches in isolation.  Finally, we introduce a scene similarity metric that correlates with human judgments.

\section{Task Description}

In the text to 3D scene generation task, the input is a natural language description, and the output is a 3D representation of a plausible scene that fits the description and can be viewed and rendered from multiple perspectives.  More precisely, given an utterance $x$ as input, the output is a scene $y$: an arrangement of 3D models representing objects at specified positions and orientations in space.

In this paper, we focus on the subproblem of lexical grounding of textual terms to 3D model referents (i.e., choosing 3D models that represent objects referred to by terms in the input utterance $x$). We employ an intermediate \emph{scene template} representation parsed from the input text to capture the physical objects present in a scene and constraints between them.  This representation is then used to generate a 3D scene (Figure~\ref{fig:text2scene-pipeline}).

A na\"ive approach to scene generation might use keyword search to retrieve 3D models.  However, such an approach is unlikely to generalize well in that it fails to capture important object attributes and spatial relations.  In order for the generated scene to accurately reflect the input description, a deep understanding of language describing environments is necessary. Many challenging subproblems need to be tackled: physical object mention detection, estimation of object attributes such as size, extraction of spatial constraints, and placement of objects at appropriate relative positions and orientations. The subproblem of lexical grounding to 3D models has a larged impact on the quality of generated scenes, as later stages of scene generation rely on having a correctly chosen set of objects to arrange.

Another challenge is that much common knowledge about the physical properties of objects and the structure of environments is rarely mentioned in natural language (e.g., that most tables are supported on the floor and in an upright orientation). Unfortunately, common 3D representations of objects and scenes used in computer graphics specify only geometry and appearance, and rarely include such information.  Prior work in text to 3D scene generation focused on collecting manual annotations of object properties and relations~\cite{rouhizadeh2011collecting,coyne2012annotation}, which are used to drive rule-based generation systems.  Regrettably, the task of scene generation has not yet benefited from recent related work in NLP.

\begin{figure*}
  \includegraphics[width=\linewidth]{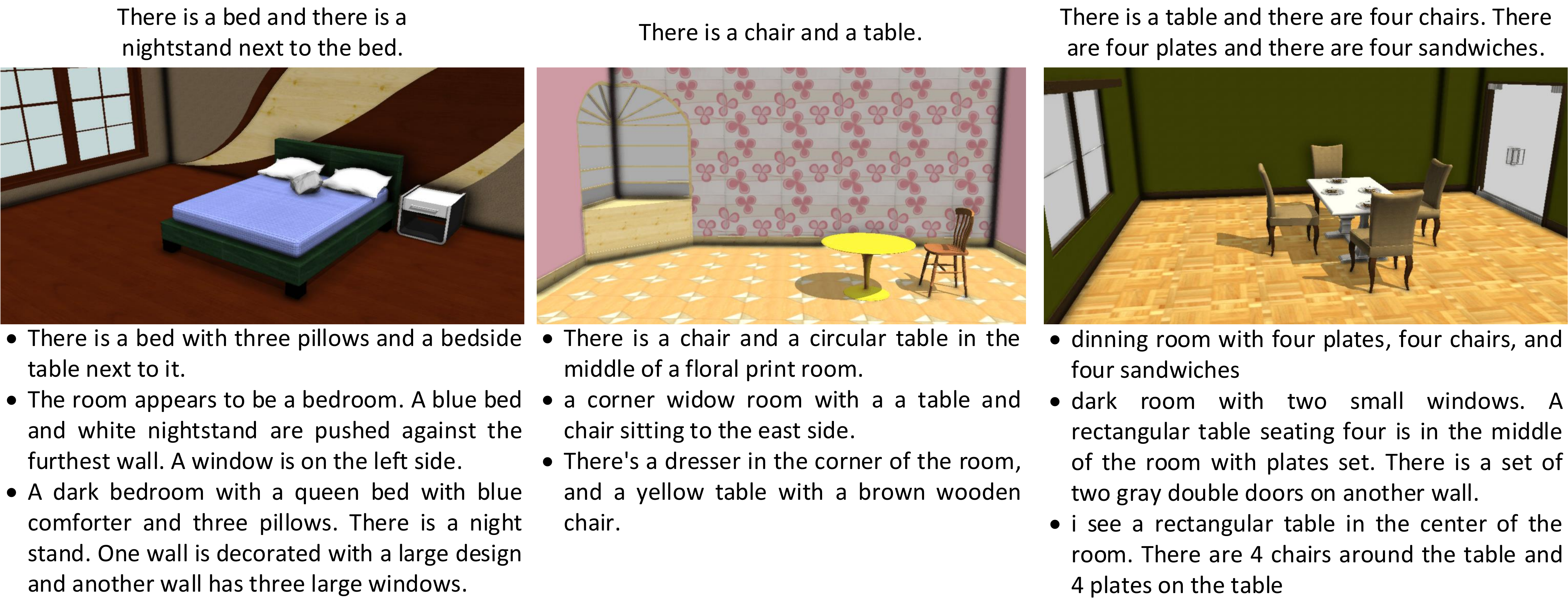} %
  \caption{Scenes created by participants from seed description sentences (\textbf{top}). Additional descriptions provided by other participants from the created scene (\textbf{bottom}). Our dataset contains around 19 scenes per seed sentence, for a total of 1129 scenes.  Scenes exhibit variation in the specific objects chosen and their placement. Each scene is described by 3 or 4 other people, for a total of 4358 descriptions.}
  \label{fig:sceneExample}
\end{figure*}

\section{Related Work}

There is much prior work in image retrieval given textual queries; a recent overview is provided by~\newcite{siddiquie2011image}.  The image retrieval task bears some similarity to our task insofar as 3D scene retrieval is an approach that can approximate 3D scene generation.

However, there are fundamental differences between 2D images and 3D scenes. Generation in image space has predominantly focused on composition of simple 2D clip art elements, as exemplified recently by~\newcite{zitnick2013learning}.  The task of composing 3D scenes presents a much higher-dimensional search space of scene configurations where finding plausible and desirable configurations is difficult.  Unlike prior work in clip art generation which uses a small pre-specified set of objects, we ground to a large database of objects that can occur in various indoor environments: 12490 3D models from roughly 270 categories.

\subsection{Text to Scene Systems}

Pioneering work on the SHRDLU system~\cite{winograd1972understanding} demonstrated linguistic manipulation of objects in 3D scenes.  However, the discourse domain was restricted to a micro-world with simple geometric shapes to simplify parsing and grounding of natural language input.  More recently, prototype text to 3D scene generation systems have been built for broader domains, most notably the \mbox{WordsEye} system~\cite{coyne2001wordseye} and later work by~\newcite{seversky2006real}. \newcite{chang2014spatial} showed it is possible to learn spatial priors for objects and relations directly from 3D scene data.

These systems use manually defined mappings between language and their representation of the physical world.  This prevents generalization to more complex object descriptions, variations in word choice and spelling, and other languages.  It also forces users to use unnatural language to express their intent (e.g., \lang{the table is two feet to the south of the window}).

We propose reducing reliance on manual lexicons by learning to map descriptions to objects from a corpus of 3D scenes and associated textual descriptions. While we find that lexical knowledge alone is not sufficient to generate high-quality scenes, a learned approach to lexical grounding can be used in combination with a rule-based system for handling compositional knowledge, resulting in better scenes than either component alone.

\subsection{Related Tasks}

Prior work has generated sentences that describe 2D images~\cite{farhadi2010every,kulkarni2011baby,karpathy2014deepbi} and referring expressions for specific objects in images~\cite{fitzgerald2013learning,kazemzadeh2014referitgame}. %
However, generating scenes is currently out of reach for purely image-based approaches. 3D scene representations serve as an intermediate level of structure between raw image pixels and simpler microcosms (e.g., grid and block worlds). This level of structure is amenable to the generation task but still realistic enough to present a variety of challenges associated with natural scenes.

A related line of work focuses on grounding referring expressions to referents in 3D worlds with simple colored geometric shapes~\cite{gorniak2004grounded,gorniak2005probabilistic}.  More recent work grounds text to object attributes such as color and shape in images~\cite{matuszek2012joint,krishnamurthy2013jointly}. \newcite{golland2010game} ground spatial relationship language in 3D scenes (e.g., \lang{to the left of}, \lang{behind}) by learning from pairwise object relations provided by crowd-workers. In contrast, we ground general descriptions to a wide variety of possible objects. The objects in our scenes represent a broader space of possible referents than the first two lines of work. Unlike the latter work, our descriptions are provided as unrestricted free-form text, rather than filling in specific templates of object references and fixed spatial relationships.

\section{Dataset}

We introduce a new dataset of 1128 scenes and 4284 free-form natural language descriptions of these scenes.\footnote{Available at \url{http://nlp.stanford.edu/data/text2scene.shtml}.} To create this training set, we used a simple online scene design interface that allows users to assemble scenes using available 3D models of common household objects (each model is annotated with a category label and has a unique ID). We used a set of 60 seed sentences describing simple configurations of interior scenes as prompts and asked workers on the Amazon Mechanical Turk crowdsourcing platform to create scenes corresponding to these seed descriptions.  To obtain more varied descriptions for each scene, we asked other workers to describe each scene.  Figure~\ref{fig:sceneExample} shows examples of seed description sentences, 3D scenes created by people given those descriptions, and new descriptions provided by others viewing the created scenes.

We manually examined a random subset of the descriptions (approximately 10\%) to eliminate spam and unacceptably poor descriptions. When we identified an unacceptable description, we also examined all other descriptions by the same worker, as most poor descriptions came from a small number of workers. From our sample, we estimate that less than 3\% of descriptions were spam or unacceptably incoherent. To reflect natural use, we retained minor typographical and grammatical errors.

Despite the small set of seed sentences, the Turker-provided scenes exhibit much variety in the specific objects used and their placements within the scene. Over 600 distinct 3D models appear in at least one scene, and more than 40\% of non-room objects are rotated from their default orientation, despite the fact that this requires an extra manipulation in the scene-building interface. The descriptions collected for these scenes are similarly diverse and usually differ substantially in length and content from the seed sentences.\footnote{Mean 26.2 words, SD 17.4; versus mean 16.6, SD 7.2 for the seed sentences. If one considers seed sentences to be the ``reference,'' the macro-averaged BLEU score \cite{papineni2002bleu} of the Turker descriptions is 12.0.}

\section{Model}

To create a model for generating scene templates from text, we train a classifier to learn lexical groundings.  We then combine our learned lexical groundings with a rule-based scene generation model.  The learned groundings allow us to select better models, while the rule-based model offers simple compositionality for handling coreference and relationships between objects. %

\subsection{Learning lexical groundings\label{sec:learned}}

To learn lexical mappings from examples, we train a classifier on a related grounding task and extract the weights of lexical features for use in scene generation.  This classifier learns from a ``discrimination'' version of our scene dataset, in which the scene in each scene--description pair is hidden among four other distractor scenes sampled uniformly at random. The training objective is to maximize the L\textsubscript{2}-regularized
log likelihood of this scene discrimination dataset under a one-vs.-all logistic regression model, using each true scene and each distractor
scene as one example (with \textit{true}/\textit{distractor} as the output label).

The learned model uses binary-valued features indicating the co-occurrence of a unigram or bigram and an object category or model ID. For example, features extracted from the scene-description pair shown in Figure~\ref{fig:text2scene-pipeline} would include the tuples $(\textit{desk}, \texttt{modelId:132})$ and $(\textit{the notepad}, \texttt{category:notepad})$.

To evaluate our learned model's performance at discriminating scenes, independently of its use in scene generation, we split our scene and description corpus (augmented with distractor scenes) randomly into train, development, and test portions 70\%-15\%-15\% by scene. Using only model ID features, the classifier achieves a discrimination accuracy of 0.715 on the test set; adding features that use object categories as well as model IDs improves accuracy to 0.833.

\subsection{Rule-based Model}

We use the rule-based parsing component described in \newcite{chang2014spatial}. This system incorporates knowledge that is important for scene generation and not addressed by our learned model (e.g., spatial relationships and coreference). In Section~\ref{sec:combined}, we describe how we use our learned model to augment this model.

\begin{figure*}
  \includegraphics[width=\textwidth]{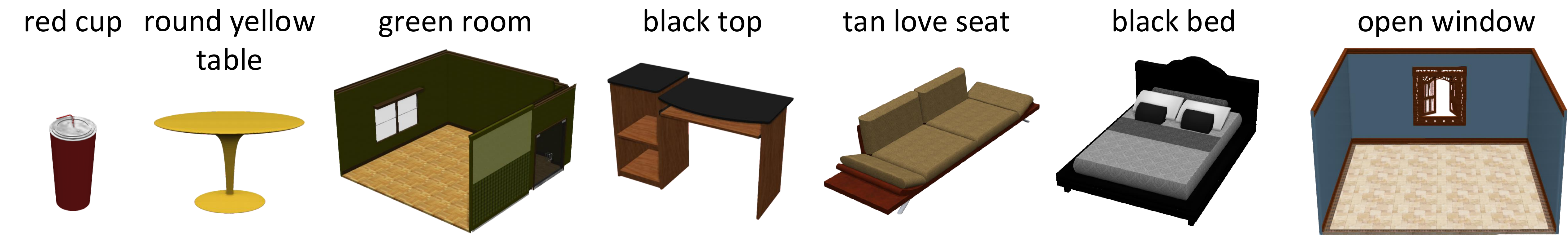}
  \caption{Some examples extracted from the top 20 highest-weight features in our learned model: lexical terms from the descriptions in our scene corpus are grounded to 3D models within the scene corpus.}
  \label{fig:grounding-examples}
\end{figure*}

This rule-based approach is a three-stage process using established NLP systems: 1) The input text is split into multiple sentences and parsed using the Stanford CoreNLP pipeline~\cite{manningstanford}. Head words of noun phrases are identified as candidate object categories, filtered using WordNet~\cite{miller1995wordnet} to only include physical objects.  2) References to the same object are collapsed using the Stanford coreference system.  3) Properties are attached to each object by extracting other adjectives and nouns in the noun phrase. These properties are later used to query the 3D model database.

We use the same model database as~\newcite{chang2014spatial} and also extract spatial relations between objects using the same set of dependency patterns.

\subsection{Combined Model\label{sec:combined}}

The rule-based parsing model is limited in its ability to choose appropriate 3D models. We integrate our learned lexical groundings with this model to build an improved scene generation system.

\paragraph{Identifying object categories}
Using the rule-based model, we extract all noun phrases as potential objects. For each noun phrase $p$, we extract features $\{\phi_i\}$ and compute the score of a category $c$ being described by the noun phrase as the sum of the feature weights from the learned model in Section~\ref{sec:learned}:
$$
\operatorname{Score}(c \mid p) = \sum_{\phi_i \in \phi(p)} \theta_{(i,c)},
$$ 
where $\theta_{(i,c)}$ is the weight for associating feature $\phi_i$ with category $c$. From categories with a score higher than $T_c = 0.5$, we select the best-scoring category as the representative for the noun phrase.  If no category's score exceeds $T_c$, we use the head of the noun phrase for the object category.

\paragraph{3D model selection} For each object mention detected in the description, we use the feature weights from the learned model to select a specific object to add to the scene.  After using dependency rules to extract spatial relationships and descriptive terms associated with the object, we compute the score of a 3D model $m$ given the category $c$ and a set of descriptive terms $d$ using a similar sum of feature weights. As the rule-based system may not accurately identify the correct set of terms $d$, we augment the score with a sum of feature weights over the entire input description $x$:
$$
m = \argmax_{m\in\{c\}} \lambda_d \sum_{\phi_i \in \phi(d)}\theta_{(i,m)} + \lambda_x \sum_{\phi_i \in \phi(x)} \theta_{(i,m)}
$$
For the results shown here, $\lambda_d = 0.75$ and $\lambda_x = 0.25$. We select the best-scoring 3D model with positive score.  If no model has positive score, we assume the object mention was spurious and omit the object.

\begin{table}
  \centering\footnotesize
  \begin{tabular}{ll@{\hskip 2em}ll}
    \toprule
    text   & category          & text   & category \\
    \midrule
    chair  & \cat{Chair}       & round  & \cat{RoundTable} \\
    lamp   & \cat{Lamp}        & laptop & \cat{Laptop} \\
    couch  & \cat{Couch}       & fruit & \cat{Bowl} \\
    vase   & \cat{Vase}        & round table & \cat{RoundTable} \\
    sofa   & \cat{Couch}       & laptop & \cat{Computer} \\
    bed    & \cat{Bed}         & bookshelf & \cat{Bookcase} \\
    \bottomrule
  \end{tabular}
  \caption{Top groundings of lexical terms in our dataset to categories of 3D models in the scenes.}
  \label{tab:groundedCats}
\end{table}

\section{Learned lexical groundings}

By extracting high-weight features from our learned model, we can visualize specific models to which lexical terms are grounded (see Figure~\ref{fig:grounding-examples}). These features correspond to high frequency text--3D model pairs within the scene corpus.  Table~\ref{tab:groundedCats} shows some of the top learned lexical groundings to model database categories.  We are able to recover many simple identity mappings without using lexical similarity features, and we capture several lexical variants (e.g., \lang{sofa} for \cat{Couch}).  A few erroneous mappings reflect common co-occurrences; for example, \lang{fruit} is mapped to \cat{Bowl} due to fruit typically being observed in bowls in our dataset.

\begin{figure*}
  \includegraphics[width=\textwidth]{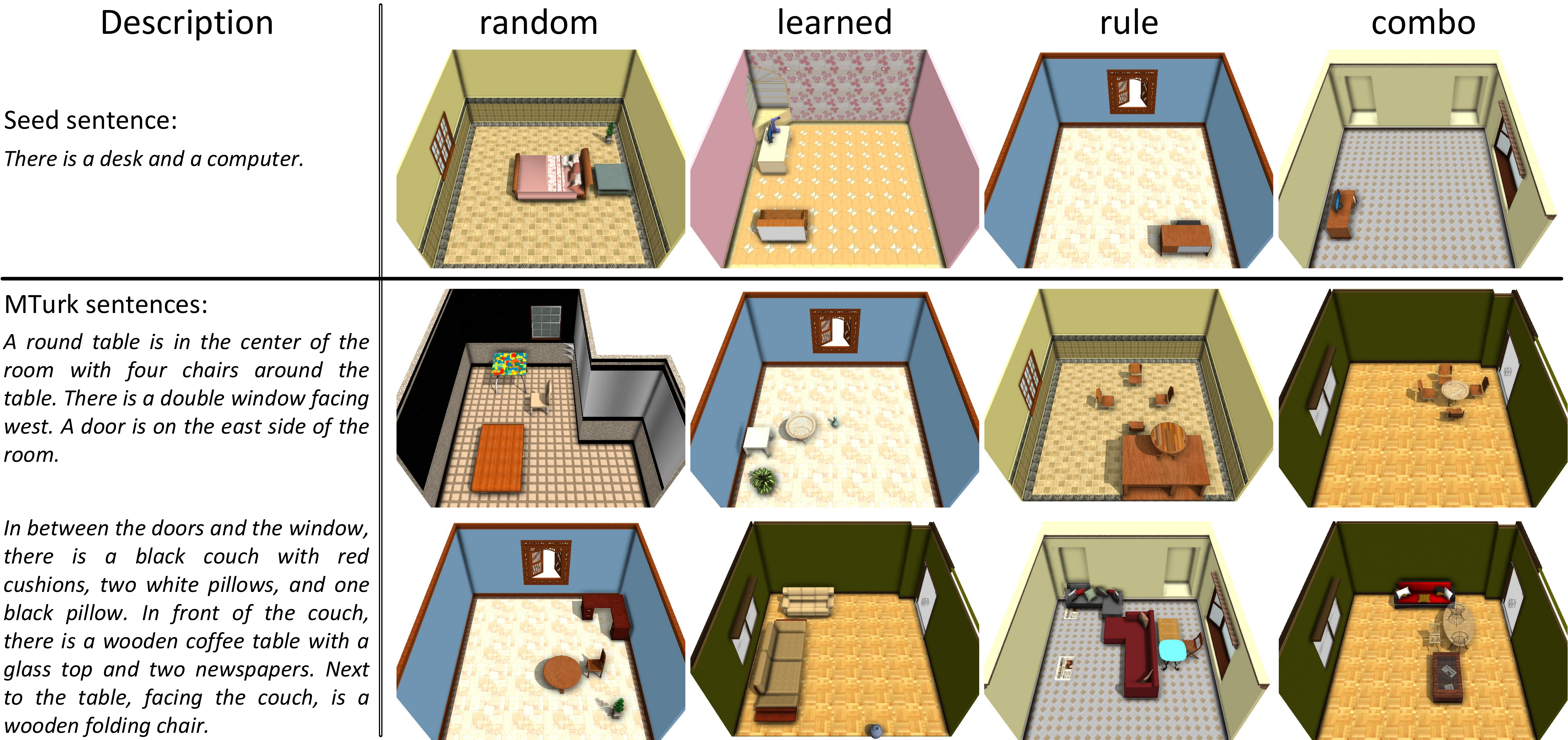} %
  \caption{Qualitative comparison of generated scenes for three input descriptions (one \emph{Seed} and two \emph{MTurk}), using the four different methods: \emph{random}, \emph{learned}, \emph{rule}, \emph{combo}.}
  \label{fig:exampleResults}
\end{figure*}

\section{Experimental Results}

We conduct a human judgment experiment to compare the quality of generated scenes using the approaches we presented and baseline methods. To evaluate whether lexical grounding improves scene generation, we need a method to arrange the chosen models into 3D scenes.  Since 3D scene layout is not a focus of our work, we use an approach based on prior work in 3D scene synthesis and text to scene generation~\cite{fisher2012example,chang2014spatial}, simplified by using sampling rather than a hill climbing strategy.

\paragraph{Conditions} We compare five conditions: \{\emph{random,  learned, rule, combo, human}\}.  The \textit{random} condition represents a baseline which synthesizes a scene with randomly-selected models, while \emph{human} represents scenes created by people. The \emph{learned} condition takes our learned lexical groundings, picks the four\footnote{The average number of objects in a scene in our human-built dataset was 3.9.} most likely objects, and generates a scene based on them. The \emph{rule} and \emph{combo} conditions use scenes generated by the rule-based approach and the combined model, respectively.

\paragraph{Descriptions} We consider two sets of input descriptions: \emph{\{Seeds, MTurk\}}.  The \emph{Seeds} descriptions are 50 of the initial seed sentences from which workers were asked to create scenes.  These seed sentences were simple (e.g., \lang{There is a desk and a chair}, \lang{There is a plate on a table}) and did not have modifiers describing the objects.  The \emph{MTurk} descriptions are much more descriptive and exhibit a wider variety in language (including misspellings and ungrammatical constructs).  Our hypothesis was that the rule-based system would perform well on the simple \emph{Seeds} descriptions, but it would be insufficient for handling the complexities of the more varied \emph{MTurk} descriptions.  For these more natural descriptions, we expected our combination model to perform better. Our experimental results confirm this hypothesis.

\subsection{Qualitative Evaluation}

Figure~\ref{fig:exampleResults} shows a qualitative comparison of 3D scenes generated from example input descriptions using each of the four methods. In the top row, the \emph{rule-based} approach selects a CPU chassis for \lang{computer}, while \emph{combo} and \emph{learned} select a more iconic monitor.  In the bottom row, the rule-based approach selects two newspapers and places them on the floor, while the combined approach correctly selects a coffee table with two newspapers on it.  The learned model is limited to four objects and does not have a notion of object identity, so it often duplicates objects.

\subsection{Human Evaluation}

We performed an experiment in which people rated the degree to which scenes match the textual descriptions from which they were generated.  Such ratings are a natural way to evaluate how well our approach can generate scenes from text: in practical use, a person would provide an input description and then judge the suitability of the resulting scenes.  For the \emph{MTurk} descriptions, we randomly sampled 100 descriptions from the development split of our dataset.

\begin{figure}
  \includegraphics[width=\columnwidth]{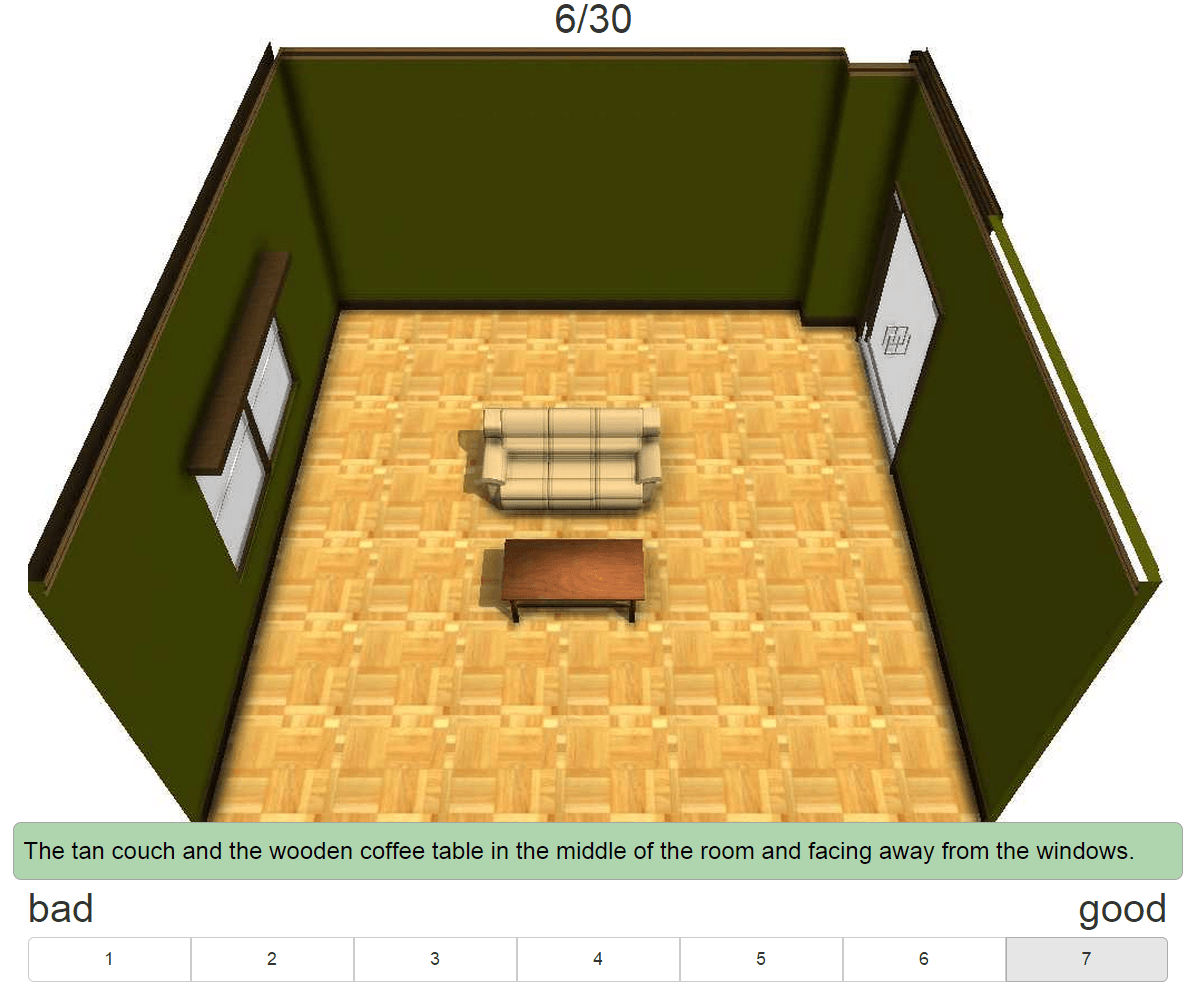}
  \caption{Screenshot of the UI for rating scene-description match.}
  \label{fig:evaluation-ui}
\end{figure}

\paragraph{Procedure}
During the experiment, each participant was shown 30 pairs of scene descriptions and generated 3D scenes drawn randomly from all five conditions.  All participants provided 30 responses each for a total of 5040 scene-description ratings.  Participants were asked to rate how well the generated scene matched the input description on a 7-point Likert scale, with 1 indicating a poor match and 7 a very good one (see Figure~\ref{fig:evaluation-ui}). In a separate task with the same experimental procedure, we asked other participants to rate the overall plausibility of each generated scene without a reference description.  This plausibility rating measures whether a method can generate plausible scenes irrespective of the degree to which the input description is matched.  We used Amazon Mechanical Turk to recruit 168 participants for rating the match of scenes to descriptions and 63 participants for rating scene plausibility.

\paragraph{Design}
The experiment followed a within-subjects factorial design.  The dependent measure was the Likert rating.  Since per-participant and per-scene variance on the rating is not accounted for by a standard ANOVA, we use a mixed effects model which can account for both fixed effects and random effects to determine the statistical significance of our results.\footnote{We used the \texttt{lme4} R package and optimized fit with maximum log-likelihood~\cite{baayen2008mixed}. We report significance results using the likelihood-ratio (LR) test.}  We treat the participant and the specific scene as random effects of varying intercept, and the method condition as the fixed effect.

\newcommand{\ci}[2]{{\footnotesize(#1\,--\,#2)}}
\begin{table}
  \centering
  \begin{tabular}{l r@{ \ }r r@{ \ }r }
    \toprule
    method & \multicolumn{2}{c}{Seeds} & \multicolumn{2}{c}{MTurk}    \\
    \midrule
    random & 2.03 &\ci{1.88}{2.18}  & 1.68 &\ci{1.57}{1.79} \\
    learned& 3.51 &\ci{3.23}{3.77}  & 2.61 &\ci{2.40}{2.84} \\
    rule   & \textbf{5.44} &\ci{5.26}{5.61}  & 3.15 &\ci{2.91}{3.40} \\
    combo  & 5.23 &\ci{4.96}{5.44}  & \textbf{3.73} &\ci{3.48}{3.95} \\[1ex]
    human  & 6.06 &\ci{5.90}{6.19}  & 5.87 &\ci{5.74}{6.00} \\
    \bottomrule
  \end{tabular}
  \caption{Average scene-description match ratings across sentence types and methods (95\% C.I.).}
  \label{tab:humanEvalResults}
\end{table}

%
%
%
%
%

\paragraph{Results}

There was a significant effect of the method condition on the scene-description match rating: \lrtest{4}{5040}{1378.2}{0.001}.  Table~\ref{tab:humanEvalResults} summarizes the average scene-description match ratings and 95\% confidence intervals for all sentence type--condition pairs.  All pairwise differences between ratings were significant under Wilcoxon rank-sum tests with the Bonferroni-Holm correction ($p\,\textless\,$0.05).  The scene plausibility ratings, which were obtained independent of descriptions, indicated that the only significant difference in plausibility was between scenes created by people (\emph{human}) and all the other conditions.  We see that for the simple seed sentences both the rule-based and combined model approach the quality of human-created scenes.  However, all methods have significantly lower ratings for the more complex \emph{MTurk} sentences. In this more challenging scenario, the combined model is closest to the manually created scenes and significantly outperforms both rule-based and learned models in isolation.

\begin{figure}
  \includegraphics[width=\columnwidth]{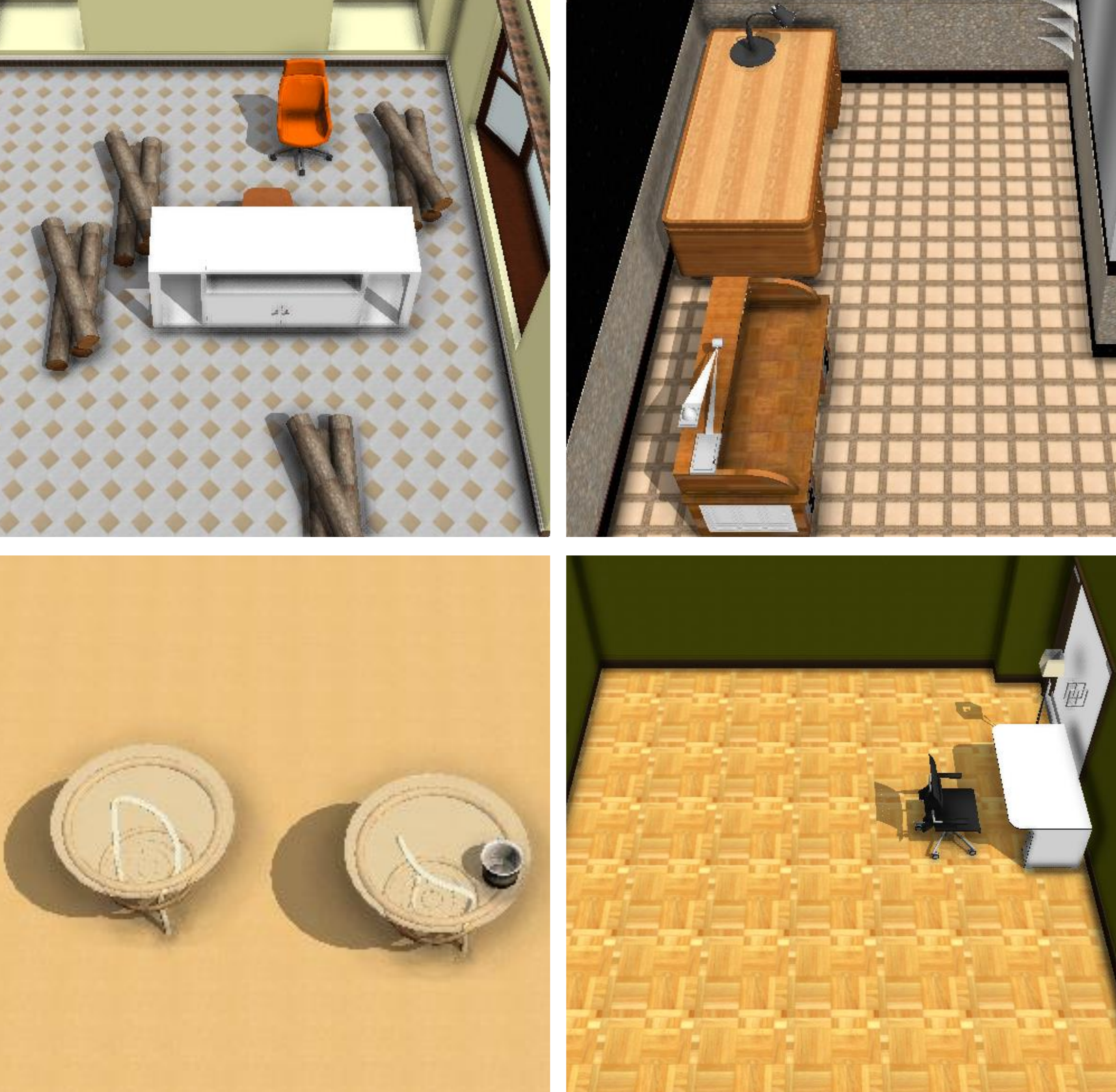} %
  \caption{Common scene generation errors. From top left clockwise: \emph{Wood table and four wood chairs in the center of the room}; \emph{There is a black and brown desk with a table lamp and flowers}; \emph{There is a white desk, a black chair, and a lamp in the corner of the room}; \emph{There in the middle is a table, on the table is a cup}.}
  \label{fig:error-wood}
\end{figure}

\subsection{Error Analysis}

Figure~\ref{fig:error-wood} shows some common error cases in our system.  The top left scene was generated with the rule-based method, the top right with the learned method, and the bottom two with the combined approach.  At the top left, there is an erroneous selection of concrete object category (wood logs) for the \lang{four wood chairs} reference in the input description, due to an incorrect head identification.  At top right,  the learned model identifies the presence of brown desk and lamp but erroneously picks two desks and two lamps (since we always pick the top four objects).  The scene on the bottom right does not obey the expressed spatial constraints (\lang{in the corner of the room}) since our system does not understand the grounding of room corner and that the top right side is not a good fit due to the door.  In the bottom left, incorrect coreference resolution results in two tables for \lang{There in the middle is a table, on the table is a cup.}

\subsection{Scene Similarity Metric}

We introduce an automated metric for scoring scenes given a scene template representation, the \term{aligned scene template similarity} (ASTS). Given a one-to-one alignment $A$ between the nodes of a scene template and the objects in a scene, let the alignment penalty $J(A)$ be the sum of the number of unaligned nodes in the scene template and the number of unaligned objects in the scene.  For the aligned nodes, we compute a similarity score $S$ per node pair $(n,n')$ where $S(n,n') = 1$ if the model ID matches, $S(n,n') = 0.5$ if only the category matches and 0 otherwise.

We define the ASTS of a scene with respect to a scene template to be the maximum alignment score over all such alignments:
\[\operatorname{ASTS}(s, z) = \max_A \frac{\sum_{(n,n') \in A} S(n,n')}{J(A)+|A|}.\]

With this definition, we compare average ASTS scores for each method against average human ratings (Table~\ref{tab:sceneSimResults}). We test the correlation of the ASTS metric against human ratings using Pearson's $r$ and Kendall's rank correlation coefficient $r_\tau$. %
We find that ASTS and human ratings are strongly correlated ($r=0.70$, $r_\tau=0.49$, $p < 0.001$).  This suggests ASTS scores could be used to train and algorithmically evaluate scene generation systems that map descriptions to scene templates.

\begin{table}
  \centering
  \begin{tabular}{lcc}
    \toprule
    method & Human & ASTS \\
    \midrule
    random & 1.68  & 0.08 \\
    learned& 2.61  & 0.23 \\
    rule   & 3.15 & 0.32 \\
    combo  & \textbf{3.73}  & \textbf{0.44} \\
    \bottomrule& 
  \end{tabular}
  \caption{Average human ratings (out of 7) and aligned scene template similarity scores.}
  \label{tab:sceneSimResults}
\end{table}

\section{Future Work}

Many error cases in our generated scenes resulted from not interpreting spatial relations.  An obvious improvement would be to expand our learned lexical grounding approach to include spatial relations. This would help with spatial language that is not handled by the rule-based system's dependency patterns (e.g., \lang{around}, \lang{between}, \lang{on the east side}).  One approach would be to add spatial constraints to the definition of our scene similarity score and use this improved metric in training a semantic parser to generate scene templates.

To choose objects, our current system uses information obtained from language--object co-occurrences and sparse manually-annotated category labels; another promising avenue for achieving better lexical grounding is to propagate category labels using geometric and image features to learn the categories of unlabeled objects. Novel categories can also be extracted from Turker descriptions. These new labels could be used to improve the annotations in our 3D model database, enabling a wider range of object types to be used in scene generation.

Our approach learns object references without using lexical similarity features or a manually-assembled lexicon. Thus, we expect that our method for lexical grounding can facilitate development of text-to-scene systems in other languages. However, additional data collection and experiments are necessary to confirm this and identify challenges specific to other languages.

The necessity of handling omitted information suggests that a model incorporating a more sophisticated theory of pragmatic inference could be beneficial. Another important problem not addressed here is the role of context and discourse in interpreting scene descriptions.  For example, several of our collected descriptions include language imagining embodied presence in the scene (e.g., \lang{The wooden table is to your right, if you're entering the room from the doors}).

\section{Conclusion}

Prior work in 3D scene generation relies on purely rule-based methods to map object references to concrete 3D objects.  We introduce a dataset of 3D scenes annotated with natural language descriptions which we believe will be of great interest to the research community.  Using this corpus of scenes and descriptions, we present an approach that learns from data how to ground textual descriptions to objects.

To evaluate how our grounding approach impacts generated scene quality, we collect human judgments of generated scenes. In addition, we present a metric for automatically comparing generated scene templates to scenes, and we show that it correlates strongly with human judgments.

We demonstrate that rich lexical grounding can be learned directly from an unaligned corpus of 3D scenes and natural language descriptions, and that our model can successfully ground lexical terms to concrete referents, improving scene generation over baselines adapted from prior work.

\section*{Acknowledgments}
We thank Katherine Breeden for valuable feedback. The authors gratefully acknowledge the support of the Defense Advanced Research Projects Agency (DARPA) Deep Exploration and Filtering of Text (DEFT) Program under Air Force Research Laboratory (AFRL) contract no.\ FA8750-13-2-0040, the National Science Foundation under grant no.\ IIS 1159679, the Department of the Navy, Office of Naval Research, under grant no.\ N00014-10-1-0109, and the Stanford Graduate Fellowship fund. Any opinions, findings, and conclusions or recommendations expressed in this material are those of the authors and do not necessarily reflect the views of the National Science Foundation, the Office of Naval Research, DARPA, AFRL, or the US government.

\bibliographystyle{acl}
\bibliography{lexground}

\begin{thebibliography}{}

\bibitem[\protect\citename{Baayen \bgroup et al.\egroup }2008]{baayen2008mixed}
R.H. Baayen, D.J. Davidson, and D.M. Bates.
\newblock 2008.
\newblock Mixed-effects modeling with crossed random effects for subjects and
  items.
\newblock {\em Journal of Memory and Language}, 59(4):390--412.

\bibitem[\protect\citename{Chang \bgroup et al.\egroup }2014]{chang2014spatial}
Angel~X. Chang, Manolis Savva, and Christopher~D. Manning.
\newblock 2014.
\newblock Learning spatial knowledge for text to {3D} scene generation.
\newblock In {\em Proceedings of Empirical Methods in Natural Language
  Processing (EMNLP)}.

\bibitem[\protect\citename{Coyne and Sproat}2001]{coyne2001wordseye}
Bob Coyne and Richard Sproat.
\newblock 2001.
\newblock {WordsEye}: an automatic text-to-scene conversion system.
\newblock In {\em Proceedings of the 28th Annual Conference on Computer
  Graphics and Interactive Techniques}.

\bibitem[\protect\citename{Coyne \bgroup et al.\egroup
  }2012]{coyne2012annotation}
Bob Coyne, Alexander Klapheke, Masoud Rouhizadeh, Richard Sproat, and Daniel
  Bauer.
\newblock 2012.
\newblock Annotation tools and knowledge representation for a text-to-scene
  system.
\newblock {\em Proceedings of COLING 2012: Technical Papers}.

\bibitem[\protect\citename{Farhadi \bgroup et al.\egroup
  }2010]{farhadi2010every}
Ali Farhadi, Mohsen Hejrati, Mohammad~Amin Sadeghi, Peter Young, Cyrus
  Rashtchian, Julia Hockenmaier, and David Forsyth.
\newblock 2010.
\newblock Every picture tells a story: Generating sentences from images.
\newblock In {\em Computer Vision--ECCV 2010}.

\bibitem[\protect\citename{Fisher \bgroup et al.\egroup
  }2012]{fisher2012example}
Matthew Fisher, Daniel Ritchie, Manolis Savva, Thomas Funkhouser, and Pat
  Hanrahan.
\newblock 2012.
\newblock Example-based synthesis of {3D} object arrangements.
\newblock {\em ACM Transactions on Graphics (TOG)}, 31(6):135.

\bibitem[\protect\citename{FitzGerald \bgroup et al.\egroup
  }2013]{fitzgerald2013learning}
Nicholas FitzGerald, Yoav Artzi, and Luke Zettlemoyer.
\newblock 2013.
\newblock Learning distributions over logical forms for referring expression
  generation.
\newblock In {\em Proceedings of Empirical Methods in Natural Language
  Processing (EMNLP)}.

\bibitem[\protect\citename{Golland \bgroup et al.\egroup
  }2010]{golland2010game}
Dave Golland, Percy Liang, and Dan Klein.
\newblock 2010.
\newblock A game-theoretic approach to generating spatial descriptions.
\newblock In {\em Proceedings of Empirical Methods in Natural Language
  Processing (EMNLP)}.

\bibitem[\protect\citename{Gorniak and Roy}2004]{gorniak2004grounded}
Peter Gorniak and Deb Roy.
\newblock 2004.
\newblock Grounded semantic composition for visual scenes.
\newblock {\em Journal of Artificial Intelligence Research (JAIR)},
  21(1):429--470.

\bibitem[\protect\citename{Gorniak and Roy}2005]{gorniak2005probabilistic}
Peter Gorniak and Deb Roy.
\newblock 2005.
\newblock Probabilistic grounding of situated speech using plan recognition and
  reference resolution.
\newblock In {\em Proceedings of the 7th International Conference on Multimodal
  Interfaces}.

\bibitem[\protect\citename{Karpathy \bgroup et al.\egroup
  }2014]{karpathy2014deepbi}
Andrej Karpathy, Armand Joulin, and Li~Fei-Fei.
\newblock 2014.
\newblock Deep fragment embeddings for bidirectional image sentence mapping.
\newblock In {\em Advances in Neural Information Processing Systems}.

\bibitem[\protect\citename{Kazemzadeh \bgroup et al.\egroup
  }2014]{kazemzadeh2014referitgame}
Sahar Kazemzadeh, Vicente Ordonez, Mark Matten, and Tamara~L. Berg.
\newblock 2014.
\newblock {ReferItGame}: Referring to objects in photographs of natural scenes.
\newblock In {\em Proceedings of Empirical Methods in Natural Language
  Processing (EMNLP)}.

\bibitem[\protect\citename{Krishnamurthy and
  Kollar}2013]{krishnamurthy2013jointly}
Jayant Krishnamurthy and Thomas Kollar.
\newblock 2013.
\newblock Jointly learning to parse and perceive: Connecting natural language
  to the physical world.
\newblock {\em Transactions of the Association for Computational Linguistics},
  1:193--206.

\bibitem[\protect\citename{Kulkarni \bgroup et al.\egroup
  }2011]{kulkarni2011baby}
Girish Kulkarni, Visruth Premraj, Sagnik Dhar, Siming Li, Yejin Choi,
  Alexander~C. Berg, and Tamara~L. Berg.
\newblock 2011.
\newblock Baby talk: Understanding and generating simple image descriptions.
\newblock In {\em IEEE Conference on Computer Vision and Pattern Recognition
  (CVPR)}.

\bibitem[\protect\citename{Manning \bgroup et al.\egroup
  }2014]{manningstanford}
Christopher~D. Manning, Mihai Surdeanu, John Bauer, Jenny Finkel, Steven~J.
  Bethard, and David McClosky.
\newblock 2014.
\newblock The {S}tanford {C}ore{NLP} natural language processing toolkit.
\newblock In {\em Proceedings of the 52nd Annual Meeting of the Association for
  Computational Linguistics: System Demonstrations}.

\bibitem[\protect\citename{Matuszek \bgroup et al.\egroup
  }2012]{matuszek2012joint}
Cynthia Matuszek, Nicholas FitzGerald, Luke Zettlemoyer, Liefeng Bo, and Dieter
  Fox.
\newblock 2012.
\newblock A joint model of language and perception for grounded attribute
  learning.
\newblock In {\em International Conference on Machine Learning (ICML)}.

\bibitem[\protect\citename{Miller}1995]{miller1995wordnet}
George~A. Miller.
\newblock 1995.
\newblock {WordNet}: {A} lexical database for {E}nglish.
\newblock {\em Communications of the ACM}, 38(11):39--41.

\bibitem[\protect\citename{Papineni \bgroup et al.\egroup
  }2002]{papineni2002bleu}
Kishore Papineni, Salim Roukos, Todd Ward, and Wei-Jing Zhu.
\newblock 2002.
\newblock {BLEU}: A method for automatic evaluation of machine translation.
\newblock In {\em Proceedings of the 40th Annual Meeting of the Association for
  Computational Linguistics}.

\bibitem[\protect\citename{Rouhizadeh \bgroup et al.\egroup
  }2011]{rouhizadeh2011collecting}
Masoud Rouhizadeh, Margit Bowler, Richard Sproat, and Bob Coyne.
\newblock 2011.
\newblock Collecting semantic data by {Mechanical} {Turk} for the lexical
  knowledge resource of a text-to-picture generating system.
\newblock In {\em Proceedings of the Ninth International Conference on
  Computational Semantics}.

\bibitem[\protect\citename{Savva \bgroup et al.\egroup }2014]{savva2014sizes}
Manolis Savva, Angel~X. Chang, Gilbert Bernstein, Christopher~D. Manning, and
  Pat Hanrahan.
\newblock 2014.
\newblock On being the right scale: Sizing large collections of {3D} models.
\newblock In {\em SIGGRAPH Asia 2014 Workshop on Indoor Scene Understanding:
  Where Graphics meets Vision}.

\bibitem[\protect\citename{Seversky and Yin}2006]{seversky2006real}
Lee~M. Seversky and Lijun Yin.
\newblock 2006.
\newblock Real-time automatic {3D} scene generation from natural language voice
  and text descriptions.
\newblock In {\em Proceedings of the 14th Annual ACM International Conference
  on Multimedia}.

\bibitem[\protect\citename{Siddiquie \bgroup et al.\egroup
  }2011]{siddiquie2011image}
Behjat Siddiquie, Rog{\'e}rio~Schmidt Feris, and Larry~S. Davis.
\newblock 2011.
\newblock Image ranking and retrieval based on multi-attribute queries.
\newblock In {\em IEEE Conference on Computer Vision and Pattern Recognition
  (CVPR)}.

\bibitem[\protect\citename{Winograd}1972]{winograd1972understanding}
Terry Winograd.
\newblock 1972.
\newblock Understanding natural language.
\newblock {\em Cognitive Psychology}, 3(1):1--191.

\bibitem[\protect\citename{Zitnick \bgroup et al.\egroup
  }2013]{zitnick2013learning}
C.~Lawrence Zitnick, Devi Parikh, and Lucy Vanderwende.
\newblock 2013.
\newblock Learning the visual interpretation of sentences.
\newblock In {\em IEEE International Conference on Computer Vision (ICCV)}.

\end{thebibliography}

\end{document}